\documentclass[conference]{IEEEtran}
\usepackage{times}
\usepackage[pdftex]{graphicx}
\usepackage{fnpos}
\usepackage{upquote}
\usepackage{float}
\usepackage{verbatim}
\usepackage{fancyhdr}
\usepackage{mathtools}
\usepackage{pifont}
\usepackage{tablefootnote}
\usepackage{amsmath}
\usepackage{setspace}
\usepackage{amssymb}
\usepackage{dirtytalk}
\usepackage{graphicx}
\usepackage{lineno}
\usepackage{amsmath}
\usepackage{tablefootnote}
\usepackage{booktabs}
\usepackage{mathabx}
\usepackage{graphicx}
\usepackage{subcaption}
\usepackage[utf8]{inputenc}
\usepackage[export]{adjustbox}
\usepackage{wrapfig}
\usepackage{multirow}
\usepackage[table,xcdraw,x11names]{xcolor}
\usepackage{color, colortbl}
\definecolor{LightCyan}{rgb}{0.88,1,1}
\definecolor{lemonchiffon}{rgb}{1.0, 0.98, 0.8}
\usepackage{pgfplots}
\pgfplotsset{compat=newest}
\usepgfplotslibrary{groupplots}
\usepgfplotslibrary{dateplot}

\newcommand{\method}{POCD}
\usepackage{makecell}
\usepackage[numbers]{natbib}
\usepackage{multicol}
\usepackage[bookmarks=true]{hyperref}
\hypersetup{
    colorlinks,
    linkcolor={red},
    citecolor={green},
    urlcolor={blue}
}
\usepackage[font={small}]{caption}

\newcommand{\customfootnotetext}[2]{{
  \renewcommand{\thefootnote}{#1}
  \footnotetext[0]{#2}}}
\newcommand\blfootnote[1]{%
  \begingroup
  \renewcommand\thefootnote{}\footnote{#1}%
  \addtocounter{footnote}{-1}%
  \endgroup
}

\begin{document}

\title{\method : Probabilistic Object-Level Change Detection and Volumetric Mapping in Semi-Static Scenes}

\author{\authorblockN{Jingxing Qian\textsuperscript{*,1},
Veronica Chatrath\textsuperscript{*,1},
Jun Yang\textsuperscript{1}, 
James Servos\textsuperscript{2}, 
Angela P. Schoellig\textsuperscript{1},
Steven L. Waslander\textsuperscript{1}}
}

\maketitle

\customfootnotetext{*}{Equal contribution}
\customfootnotetext{1}{The authors are with the University of Toronto Institute for Aerospace Studies and the University of Toronto Robotics Institute. \\Emails: \texttt{\{firstname.lastname\}@robotics.utias.utoronto.ca}}
\customfootnotetext{2}{The author is with Clearpath Robotics, Waterloo, Canada. \\
Email: \texttt{jservos@clearpath.ai}}

\newcommand{\norm}[1]{\left\lVert#1\right\rVert}
\newcommand{\defeq}{\vcentcolon=}

\begin{abstract}

Maintaining an up-to-date map to reflect recent changes in the scene is very important, particularly in situations involving repeated traversals by a robot operating in an environment over an extended period. Undetected changes may cause a deterioration in map quality, leading to poor localization, inefficient operations, and lost robots. Volumetric methods, such as truncated signed distance functions (TSDFs), have quickly gained traction due to their real-time production of a dense and detailed map, though map updating in scenes that change over time remains a challenge. We propose a framework that introduces a novel probabilistic object state representation to track object pose changes in semi-static scenes. The representation jointly models a stationarity score and a TSDF change measure for each object. A Bayesian update rule that incorporates both geometric and semantic information is derived to achieve consistent online map maintenance. To extensively evaluate our approach alongside the state-of-the-art, we release a novel real-world dataset in a warehouse environment. We also evaluate on the public ToyCar dataset. Our method outperforms state-of-the-art methods on the reconstruction quality of semi-static environments.\blfootnote{This work was supported by the Vector Institute for Artificial Intelligence in Toronto and the NSERC Canadian Robotics Network (NCRN). \\ \indent Dataset download and Supplementary Material are available at\\ \href{https://github.com/Viky397/TorWICDataset}{https://github.com/Viky397/TorWICDataset}}

\end{abstract}

\IEEEpeerreviewmaketitle

\section{Introduction}

In recent years autonomous robots have been deployed in challenging environments for extended operations, such as warehouse robotics, self-driving, and indoor monitoring. This highlights the importance of reliable map maintenance, as an up-to-date map allows robots to navigate efficiently and interact with objects in the scene. During long-term task completion, robots may encounter dynamic objects (people, robots), moderately dynamic objects (boxes, pallets), and objects that are usually static but can be modified occasionally (fences, walls). To map these environments, many widely adopted methods make use of dense, volumetric representations such as truncated signed distance functions (TSDFs) \cite{fehr2017tsdf, voxblox, kintinuous}. However, such methods assume an independent and identically distributed (i.i.d.) occupancy in each voxel, making them prone to corruption when the scene changes. 

Recent works that attempt to handle scene dynamics extend monolithic map representations by using neural networks to segment the scene and identify potentially dynamic objects. However, most of these approaches focus primarily on either static~\cite{mur2015orb, Rosinol20icra-Kimera, whelan2016elasticfusion} or short-term dynamic situations~\cite{fehr2017tsdf,Rnz2018MaskFusionRR,dynslam, strecke2019_emfusion, Xu2019MIDFusionOO}. Promising results on table-top dynamic scenes have been demonstrated, but large, real-world environments wherein objects appear, disappear, or shift between robot traversals are overlooked. Such unobserved changes to objects pose a unique challenge to existing mapping systems, as commonly used change detection methods rely on motion tracking to identify dynamic objects, instead of assessing the consistency of a prior object pose with current sensor measurements. As a result, changes in the scene that occur when the robot is offline may not be detected easily, causing it to lose localization as the map accuracy degrades over time. \\ 
\begin{figure}[t!]
  \centering
  \includegraphics[width=\columnwidth]{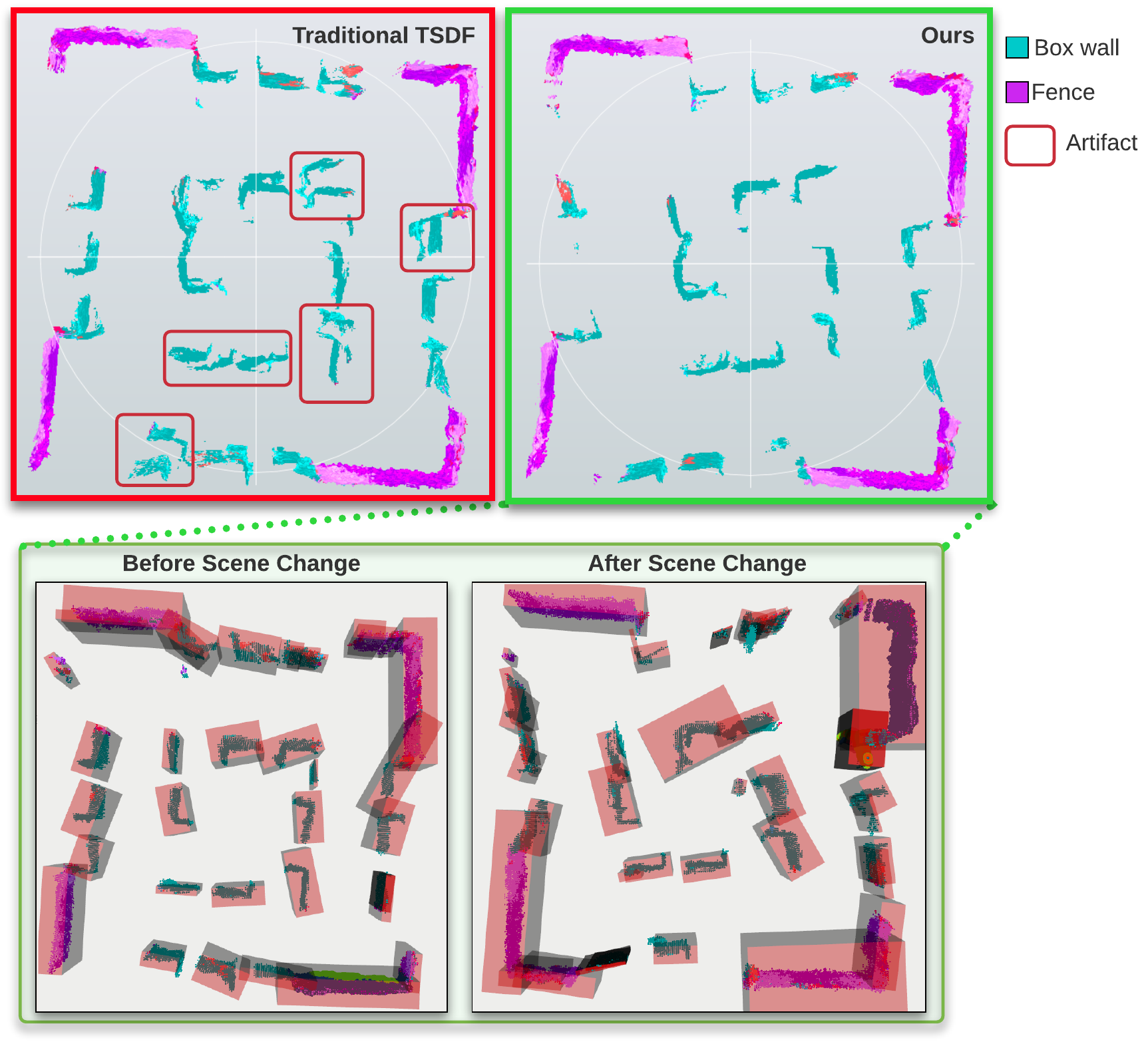}
  \caption{A comparison of semi-static scene reconstruction between our method, \method, and that of a traditional TSDF method \cite{Rosinol20icra-Kimera} on the TorWIC\_1-5 route. Using the traditional TSDF update method, artifacts, as boxed in red, persist after the box walls have moved between traversals. By comparison, our method produces a clean map that reflects the up-to-date scenario despite changed object poses. The bottom row shows our reconstructed map, along with object bounding boxes, before and after the scene has changed. The reference desired reconstruction is shown in Figure \ref{fig:results_15}.}
  \label{fig:fig1}
  \vspace{-3mm}
\end{figure}
\indent In this work, we aim to address the long-term map maintenance challenge in semi-static environments. We define as semi-static those environments that are primarily static but exhibit changed object locations between robot traversals. The primary motivation is that many useful operating environments for robots change slowly and at an object level, and these objects tend to either shift slightly, move significantly, appear or disappear. To address the impact of these changes on the accuracy of the map, we introduce a novel framework, \method, to track and correct object-level changes in the environment. The core of \method~is an object-level, probabilistic state representation that jointly models the stationarity (whether or not the object pose has changed) and the magnitude of geometric change for each object in the scene. A Bayesian update rule is derived to propagate each object's state representation by leveraging both geometric and semantic measurements. Additionally, we notice current change detection datasets focus on either table-top or household scenarios \cite{Wald2019RIO}. There exists a lack of datasets for long-term mapping in large industrial semi-static environments, such as warehouses and retail stores. With the help of Clearpath Robotics, we present the Toronto Warehouse Incremental Change Dataset (TorWIC), a representative change detection dataset in a warehouse setting. 

Our proposed method is evaluated on both the aforementioned dataset and the public ToyCar dataset~\cite{cofusion}, and is compared against state-of-the-art approaches \cite{fehr2017tsdf, Rosinol20icra-Kimera, Rnz2018MaskFusionRR, panoptictsdf}. We show that our framework produces more consistent maps despite the scene changing significantly, and is robust to sensor noise, partial observations and dynamic objects. We summarize our work with the following contributions:
\begin{itemize}
    \item We introduce a Bayesian object-level update rule with a stationarity measure and a geometric change magnitude assessment that are modelled via a joint probability distribution to detect object pose changes. The update rule leverages both geometric and semantic measurements for robust change detection in incrementally evolving scenes.
    
    \item We introduce a novel method to measure object-level geometric change using the sum of voxel differences between TSDFs.

    \item We design \method, a novel, online, object-aware mapping framework that simultaneously tracks and reconstructs changes in indoor semi-static environments which employs our Bayesian object-level update rule and geometric change calculation to propagate detected changes across entire objects, consistently. 
    
    \item We release a real-world change detection dataset captured in a warehouse setting. The environment contains static objects that change between runs, as seen by both a RGB-D camera and 2D LiDAR, with robot poses provided.
\end{itemize}

To the best of our knowledge, \method ~is the first to achieve online, probabilistic, object-level change detection and map updating for large, indoor, semi-static environments.

\section{Related Works} \label{lit_review}

\subsection{Map Representation}
The main map representations to model environments are:
\begin{itemize}
    \item sparse, feature-based methods used in vision-based systems, such as ORB ~\cite{mur2015orb,rgbdslam} and SURF features~\cite{vtar},
    \item surface representations like meshes and surfels~\cite{surfel},
    \item point cloud representations~\cite{lsd_slam}, and
    \item dense, volumetric methods such as occupancy grids~\cite{hornung13auro} and TSDFs\cite{newcombe2011kinectfusion}.   
\end{itemize}  

In particular, TSDFs have gained traction in recent years due to their unique advantage of containing rich geometric information, resulting in higher robustness to sensor noise and smoother scene reconstruction~\cite{newcombe2011kinectfusion}. Moreover, dense methods allow for measurement integration to be parallelized, resulting in efficient real-time updates.

However, most volumetric representations only capture the full geometry of the scene, ignoring object-level information. This renders them inefficient when large structures change pose, as all affected voxels must be updated through new observations to maintain an accurate map.


{
\renewcommand{\arraystretch}{2.0}
\setlength{\tabcolsep}{4pt}
\begin{table*}[t!]
\scriptsize
\centering
\caption{A comparison of indoor, changing scene datasets. }
{
\begin{tabular}{c|cccccccccc}
\Xhline{2\arrayrulewidth}  \Xhline{2\arrayrulewidth}
\textbf{Dataset} & \makecell{\textbf{Real}-\\\textbf{world}} & \textbf{Environment} & \makecell{\textbf{Full}\\\textbf{Depth}} & \makecell{\textbf{Camera}\\\textbf{Poses}}& \makecell{\textbf{Additional}\\\textbf{Sensors}}& \makecell{\textbf{Semantic}\\\textbf{Masks}}& \makecell{\textbf{Incremental}\\\textbf{Changes}}&\makecell{\textbf{Dynamic}\\\textbf{Objects}}& \textbf{Size}\\ \hline \hline

InteriorNet \cite{InteriorNet18} &  \ding{55}  & Household  &  \ding{51} & \ding{51} & IMU & \ding{51} & \ding{51} & \ding{55} & Million frames \\

Langer et al. \cite{langer_dataset} &  \ding{51}  & Household &  \ding{51} &  \ding{51} & \ding{55} &  \ding{51} & \ding{51} & \ding{55} & \makecell{31 trajectories,\\5 scenes} \\

Ambrus et al. \cite{Ambrus2016UnsupervisedOS} &  \ding{51}  & Office  &  \ding{51} & \ding{51} & \ding{55} & \ding{51} & \ding{51} & \ding{55} & \makecell{88 trajectories,\\8 scenes} \\

Co-Fusion \cite{cofusion} &  \makecell{2: \ding{55} \\ 3: \ding{51}} & \makecell{Table-top/ \\Household}  & \ding{51} & \ding{51} & \ding{55} & \ding{51} & \ding{55} & \ding{51} &  \makecell{5 trajectories,\\ 5 scenes}\\

3RScan \cite{Wald2019RIO} &  \ding{51}  & Household  & \ding{51} & \ding{51}& \ding{55} &\ding{51} & \ding{51} & \ding{55} & \makecell{1482 trajectories,\\478 scenes} \\

\makecell{TUM RGB-D \\Dynamic Objects\cite{sturm12iros}} &  \ding{51}  & Office & \ding{51}&\ding{51} & IMU &  \ding{55} & \ding{55} & \ding{51} & \makecell{9 trajectories,\\2 scenes}\\

OpenLORIS \cite{shi2019openlorisscene}   &  \ding{51}  & \makecell{Household/\\office/mall}   & \ding{51} & \ding{51} & \makecell{Odom/IMU/\\Fisheye/LiDAR} &\ding{55} &  \ding{51} & \ding{51} & \makecell{22 trajectories,\\5 scenes}\\

Fehr \textit{et al.} \cite{fehr2017tsdf}     &  \ding{51}  & Household  &   \ding{51} & \ding{51} & \ding{55} & \ding{55} & \ding{51} & \ding{51} & \makecell{23 trajectories,\\3 scenes}\\ 

ChangeSim \cite{park2021changesim} & \ding{55}  &   Warehouse  & \ding{55}  & \ding{51} & \ding{55}& \ding{51} &  \ding{51} & \ding{55} & \makecell{80 trajectories \\ $\sim$ 130k frames}\\

\textbf{\method ~(ours)}    &  \ding{51}  &  Warehouse  &  \ding{51} & \ding{51} & \makecell{Odom/IMU/\\2D LiDAR} & \ding{51}\footnotemark & \ding{51} & \ding{55} & \makecell{18 trajectories,\\ 18 scenes, $\sim$ 70k frames} \\ 
\hline
\end{tabular}
}
\label{tab:data_compare}
\end{table*}}

\subsection{Semantic Information and Object-level Reconstruction}
In recent years, researchers have expanded on volumetric map representations, incorporating both semantic and object-level information in map updates. Voxblox++ \cite{grinvald2019volumetric} and Kimera~\cite{Rosinol20icra-Kimera} extend \cite{voxblox} by incorporating voxel-level semantic information into the dense reconstruction. Recent works take this a step further by introducing objects into the map representation. Some methods use 3D CAD models to identify a set of pre-defined objects in the scene \cite{slam++}, while others construct unseen objects on-the-fly \cite{fusion++, dsg}. However, these methods still adopt the static world assumption, which is unrealistic for a robot deployed in the real world for extended operations.

\subsection{Handling of Dynamic Objects}
Recent works have attempted to handle dynamic changes in the environment, adopting one of two common strategies. The first is to specifically identify static structure classes and treat all potentially dynamic objects, usually extracted with an image-based semantic segmentation network such as Mask R-CNN \cite{8237584}, as outliers, ignoring them completely in localization and mapping\cite{DOT, Sun2019MovableObjectAwareVS, DS_SLAM, Ruchti2018MappingWD}. Though this method has proven effective when a small number of fast-moving objects are present, it can fail when used in large, crowded environments, as only a small number of static background structures will remain after dynamic object pruning~\cite{Tipaldi2013LifelongLI}. Moreover, many static objects from non-static object classes may be mission-critical (e.g., the robot may need to navigate to, and survey, pallets), and ignoring such objects may lead to inefficient task completion. \\ 
\indent The second approach is to track the dynamic objects using multi-object tracking (MOT) methods \cite{Rnz2018MaskFusionRR,Xu2019MIDFusionOO,Hachiuma2019DetectFusionDA, Wang2021DymSLAM4D}. MaskFusion~\cite{Rnz2018MaskFusionRR} decides whether an object is static based on motion inconsistency, and tracks dynamic objects by minimizing both geometric and photometric projective error. TSDF++~\cite{TSDF++} actively updates the poses of detected objects by registering their 3D models in the global frame using iterative closest point (ICP) between consecutive frames. MID-Fusion~\cite{Xu2019MIDFusionOO} also uses motion inconsistency to detect dynamic objects and estimates their poses via joint dense ICP and RGB tracking. However, such methods only handle dynamics that can be detected immediately in consecutive frames, rendering them ineffective under semi-static infrastructure changes, which occur over a longer time horizon. As well, when the robot is surrounded by moving objects, this method may aggressively update the object poses due to measurement noise and ambiguous association, leading to incorrect updates.

\subsection{Detecting Incremental Changes}
Detecting incremental changes in the scene is crucial to achieve long-term robot operation. Existing methods can be broadly classified into image-based or geometry-based approaches. Earlier image-based methods such as the works of Rosin \cite{Rosin1998ThresholdingFC} and Radke \textit{et al.} \cite{Radke2005ImageCD} compare the intensities of aligned images to detect pixel-level changes. They are usually sensitive to view-angle variation and illumination changes. Recent research applies deep learning to achieve more robust change detection. In the works of Zhan \textit{et al.} \cite{Zhan2017ChangeDB}, Guo \textit{et al.} \cite{Guo2018LearningTM}, and Varghese \textit{et al.} \cite{Varghese2018ChangeNetAD}, the authors use supervised learning to train models to detect pixel or patch-level differences. However, such learned image-based methods require densely annotated training samples and can still be sensitive to view angle and appearance changes.

Instead, researchers leverage 3D geometric information to tackle these challenges. Alcantarilla \textit{et al.} \cite{Alcantarilla2018StreetviewCD} warp coarsely registered camera images around dense reconstructions of the scene to mitigate view-angle differences. Fehr \textit{et al.}~\cite{fehr2017tsdf} update a TSDF map by directly calculating voxel-level differences between signed distance functions of the previous reconstruction and the new measurement, pruning voxels above an error threshold. However, directly comparing geometric information at a voxel-level is prone to localization and measurement noise. Yew \textit{et al.} \cite{Yew2021CityscaleSC} first use deep learning to perform non-rigid point cloud registration to mitigate localization drift, and then compare the registered point clouds to detect voxel-level changes. These methods still lack object-level information which can lead to semantically inconsistent map updates.
\addtocounter{footnote}{0}
\footnotetext{The included semantic masks are generated by a trained model. We also provide a human-annotated training set for fine-tuning. Please see the Supplementary Material for more details.}

Recently, there has been work to detect and handle incremental changes at an object level. Gomez \textit{et al.}~\cite{object_pose_graph} propose an offline method with an object pose-graph for mapping indoor semi-static environments. Data association and change estimation are performed offline between runs to update object persistence and poses.  Gomez \textit{et al.} model objects as cuboid bounding volumes instead of full 3D reconstructions, and only estimate class-level stationarity priors using a heuristic update rule, contrary to our estimate of object-level stationarity via Bayesian inference. Furthermore, they employ a centroid distance-based metric to detect object changes, which is difficult to obtain when only partial observations and incomplete object models are available.

Schmid \textit{et al.}~\cite{panoptictsdf} recently proposed a framework that also aims to construct and maintain a dense reconstruction in semi-static environments at an object-level. In \cite{panoptictsdf}, each object is represented as a submap in a layered TSDF, with a stationarity score propagated by a heuristic update rule. Though we both aim to solve the same problem, our method features an object-level association module. This allows us to find explicit associations between observations and tracked objects, and estimate geometrically meaningful changes in the scene. On the other hand, \cite{panoptictsdf} simply counts the number of inconsistent voxels, based on the weighted TSDF value differences, over overlapping submaps to determine if an update is required. Moreover, in our framework, object states are represented by a probability distribution and an update is performed via a Bayesian approach, allowing for more robustness against measurement and localization noise. 

\begin{figure*}[t!]
  \centering
  \includegraphics[width=0.95\linewidth]{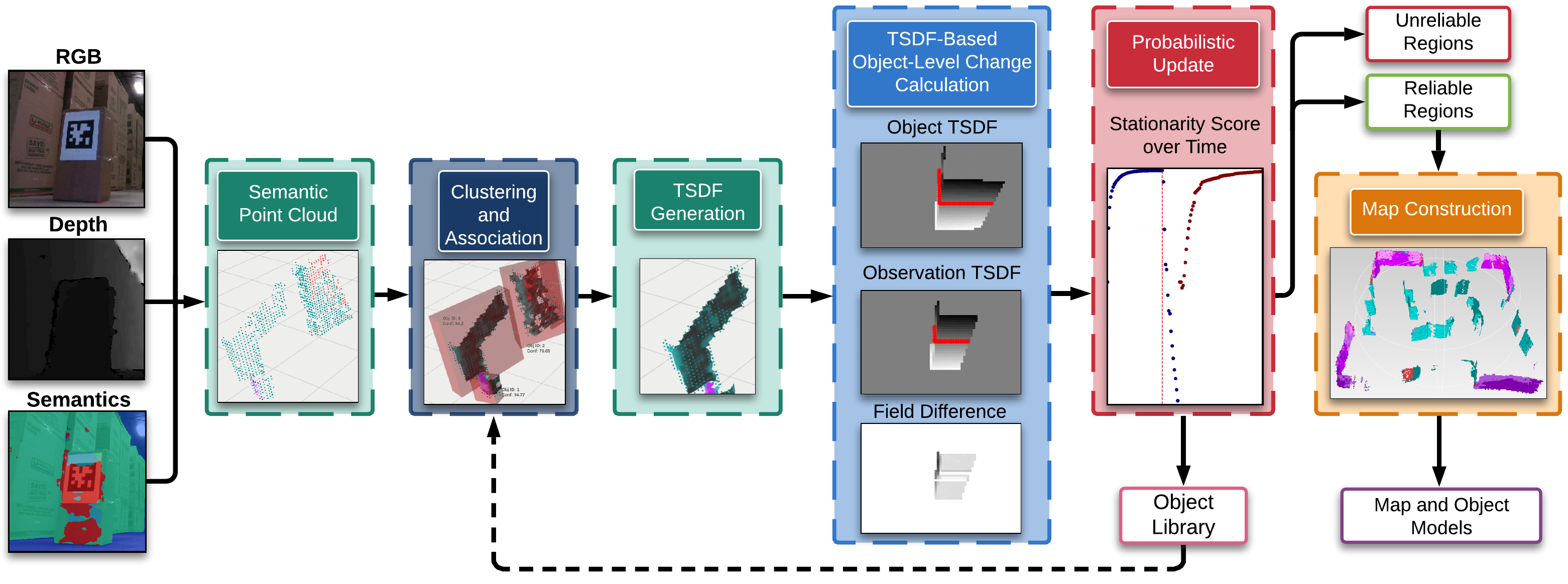}
    \caption{Our object-aware map update framework (Section~\ref{Pipeline_overview}) for semi-static environments. The system takes in RGB-D frames (Section~\ref{pipeline_representations}), each one semantically annotated and converted to a semantic point cloud (Section~\ref{sem-seg}). The point cloud is clustered into observations (Sections~\ref{3d_inst_Det}), and associated to mapped objects (Section~\ref{obj_data_assct}). A TSDF-based object-level change estimation is performed between the associated observation-object pairs, followed by a joint probabilistic update of the geometric change and stationarity score (Section~\ref{prob_stat_update}). Objects with a high stationarity score are used to generate the new map (Section~\ref{obj_map_update}), and the low-score objects are discarded.}
  \label{fig:pipeline}
\end{figure*}

\subsection{Indoor Changing Scene Datasets}
To facilitate research in long-term autonomy in indoor changing scenes, a variety of datasets have been released. The datasets, both real and simulated, are captured in different environments such as indoor offices \cite{Ambrus2016UnsupervisedOS, sturm12iros}, homes \cite{fehr2017tsdf, Wald2019RIO, cofusion, Ambrus2016UnsupervisedOS, InteriorNet18, langer_dataset}, malls \cite{shi2019openlorisscene}, and warehouses \cite{park2021changesim}. A comparison of these datasets is seen in Table~\ref{tab:data_compare}. Many of the datasets are collected using handheld devices rather than on a robot platform. The lack of additional sensors such as inertial measurement units (IMUs), odometry, and LiDAR, makes these datasets difficult to extend beyond mapping tasks (e.g., simultaneous localization and mapping (SLAM)). Moreover, a real-world dataset in a warehouse environment is lacking.

\section{System Description}

\label{Pipeline_overview}
\subsection{Overview and Assumptions}
\label{Overview_assumption}
In this work, we consider map maintenance in long-term, semi-static scenarios. Our goal is to track object-level change in the scene, reconstructing the objects when sufficient evidence of their change has been collected, to produce a final map that reflects the up-to-date configuration. In this section, we provide an overview of each part of the \method~framework. The following assumptions were made in this work:

\begin{enumerate}
    \item The operating space is a bounded, indoor environment, such as a warehouse or a retail store with rigid objects.
    \item High-level prior knowledge of the environment is available, such as object type and range of object dimensions.
    \item The poses of the robot can be obtained from an existing localization system such as in \cite{dsg}.
    \item Scene changes are due to the addition, removal, or planar motion of objects between robot traversals.
\end{enumerate}
Though the focus of this work is on semi-static environments, we also identify objects that are dynamic, to improve the overall robustness of the system. A flow diagram of the proposed framework is shown in Figure \ref{fig:pipeline}.
\vspace{-1.5mm}
\subsection{Pipeline and Representations}
\label{pipeline_representations}
We now introduce our object-aware pipeline which aims to improve long-term map quality in semi-static indoor environments. The system inputs are a sequence of colour and depth frames $\mathcal{F}=\{\mathbf{F}_t\}_{t=1 \dots T}$ taken from a RGB-D camera $\mathcal{C}$, and 6-DoF world-to-camera transformations, $\mathcal{T}^{CW}=\{\mathbf{T}^{CW}_{t}=\{\mathbf{p}^{CW}_{t},\mathbf{q}^{CW}_{t}\}\}_{t=1 \dots T}$, with 3D position $\mathbf{p}^{CW}_{t}$ and orientation $\mathbf{q}^{CW}_{t}$ at each timestep $t$, provided by an independent localization system. The framework maintains a library of mapped objects $\mathcal{O}=\{\mathbf{O}_i\}_{i=1 \dots I}$ where each object $\mathbf{O}_i$ contains: 

\begin{figure}[b!]
\vspace{-5mm}
  \centering
  \includegraphics[width=0.81\columnwidth]{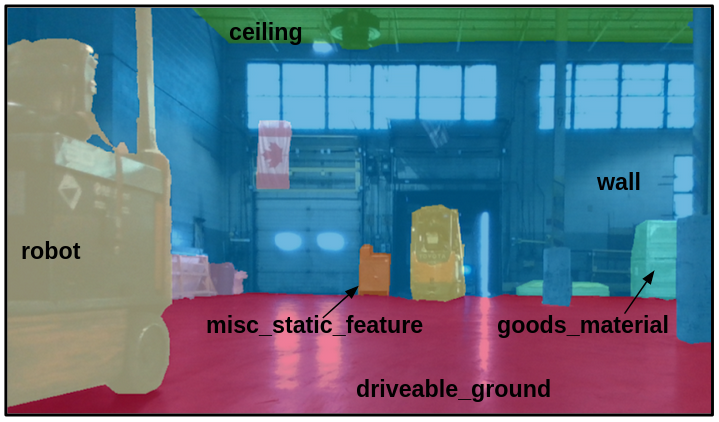}
  \caption{A sample semantically labelled training image taken in a Clearpath Robotics warehouse, with a subset of the training labels.}
  \label{fig:clearpath_labels}
  \vspace{-5mm}
\end{figure}

\begin{itemize}
    \item a 4-DoF global pose, $\mathbf{T}^{OW}_{i} = \{\mathbf{p}^{OW}_{i},\phi^{OW}_{i}\}$,
    \item a point cloud from accumulated depth data, $\mathbf{P}_{i}$, and the resulting TSDF reconstruction, $\mathbf{M}_i$, 
    \item a bounding box, $\mathbf{B}_i$, that is aligned with the major and minor axes of the object reconstruction,
    \item a semantic class, $c_i$,
    \item a state probability distribution, $p(l_i,v_i)$, to model the object-level geometric change, $l_i\in\mathbb{R}$, and the stationarity (likelihood of the object being stationary), $v_i \in [0,1]$.
\end{itemize}

The pipeline starts with an empty library. Since our experiments are conducted in a warehouse scenario, we assume the objects are restricted to planar motions. In particular, the objects can only rotate around the $z$-axis with a heading $\phi^{OW}$. The system outputs a dense, up-to-date semantic TSDF map containing the current configuration of the environment. We choose a dense reconstruction as it contains rich geometric information that can be used for many robotic tasks such as path planning, obstacle avoidance, and robot-object interaction. 

\subsection{Semantic Segmentation from RGB Images}
\label{sem-seg}
To extract semantically consistent instances from a RGB-D frame, $\mathbf{F}_{t}$, we utilize a semantic segmentation network (DeepLabV3 \cite{Chen2017RethinkingAC}) which provides pixel-level semantic predictions for 16 classes. The network was trained on a proprietary dataset of the warehouse environment provided by Clearpath Robotics, as seen in Figure \ref{fig:clearpath_labels}. 

Previous works require either instance or panoptic segmentation on their input images \cite{Rnz2018MaskFusionRR, strecke2019_emfusion, Xu2019MIDFusionOO} in order to identify unique object instances. However, obtaining instance-level segmentation is difficult in real-world warehouse and retail settings where the robot will face visually identical and cluttered objects, such as boxes and pallets, that are contiguous with one another. Therefore, we adopt a semantic segmentation method which is representative of such challenging scenes.

\subsection{3D Observations from RGB-D Frames}
\label{3d_inst_Det}

We extract a set of observations $\mathcal{Y}_{t} = \{\mathbf{Y}_{t,j}\}_{j=1 \dots J}$ from the segmented frame $\mathbf{F}_{t}$, where each observation, $\mathbf{Y}_{t,j}$, contains: 

\begin{itemize}
    \item a 4-DoF global pose, $\mathbf{T}^{YW}_{t,j} = \{\mathbf{p}^{YW}_{t,j},\phi^{YW}_{t,j}\}$,
    \item a point cloud, $\hat{\mathbf{P}}_{t,j}$, extracted from $\mathbf{F}_{t}$,
    \item a bounding box, $\mathbf{B}_{t,j}$, that is aligned with the major and minor axes of the object reconstruction,
    \item a semantic class, $c_{t,j}$,
    \item and a stationarity class, $s_{t,j}$.
\end{itemize}

To obtain $\mathcal{Y}_{t}$, we convert $\mathbf{F}_{t}$ into a semantically segmented point cloud, $\mathbf{P}^{\textrm{raw}}_{t}$, transforming it to align with the world frame based on the current camera pose, $\mathbf{T}^{CW}_{t}$. We first remove the ground-plane and reject outliers from $\mathbf{P}^{\textrm{raw}}_{t}$. Since our target environments are warehouses and retail spaces, many planar features exist in the scene. Thus, we search for planes in $\mathbf{P}^{\textrm{raw}}_{t}$, and enforce semantic consistency over the co-planar points via a majority voting scheme to reject noise from incorrect semantic segmentation. This yields a filtered point cloud, $\mathbf{P}^{\textrm{filt}}_{t}$. Note that other shape priors can be integrated based on prior knowledge of the target environments.

Using Euclidean clustering, we extract semantically consistent clusters, $\{\hat{\mathbf{P}}_{t,j}\}_{j=1 \dots J}$, from $\mathbf{P}^{\textrm{filt}}_{t}$. An observation, $\mathbf{Y}_{t,j}$, is spawned for each cluster, $\hat{\mathbf{P}}_{t,j}$. To retrieve the 4-DoF pose, $\mathbf{T}^{YW}_{t,j}$, of $\mathbf{Y}_{t,j}$, we follow a two-step approach. $z$-axis aligned 3D bounding cuboid, $\mathbf{B}_{t,j}$, is fit to $\hat{\mathbf{P}}_{t,j}$, with its $x$ and $y$ axes aligned with the major and minor axes of the flattened $\hat{\mathbf{P}}_{t,j}$ using principle component analysis (PCA), to acquire its heading, $\phi^{YW}_{t,j}$. The translation, $\mathbf{p}^{YW}_{t,j}$, is set to the centroid of $\mathbf{B}_{t,j}$. Observations with overlapping bounding cuboids are then merged, and semantic consistency is once again enforced.

A semantic class, $c_{t,j}$, is assigned to each observation, $\mathbf{Y}_{t,j}$, based on the enforced semantic prediction over the points. Finally, a stationarity class, $s_{t,j}$, can be mapped from $c_{t,j}$, where $s_{t,j}=s(c_{t,j}) \in \{0,1\}$, by leveraging prior class property knowledge. The value of $s_{t,j}=0$ denotes a dynamic object and the value of $s_{t,j}=1$ denotes a static object. For example, an object with $c_{t,j}=robot$ will have $s_{t,j}=0$ whereas an object with $c_{t,j}=shelf$ will have $s_{t,j}=1$.
\subsection{Object-Level Data Association}
\label{obj_data_assct}
The extracted observations, $\mathcal{Y}_t$, are compared to existing objects in the mapped object library, $\mathcal{O}$. Point-to-plane ICP, based on the Point Cloud Library (PCL) \cite{rusu20113d}, is first run between each object-observation pair, $\{\mathbf{O}_i,\mathbf{Y}_{t,j}\}$, with a centroid distance below a generous association distance threshold, $\theta_{\textrm{dist}}$, to find both the relative transformation, $\mathbf{T}_{ij}=\{\mathbf{p}_{ij},\phi_{ij}\}$, and geometric dissimilarity, $\epsilon_{ij}$ (percent of outliers after ICP convergence), for each pair. A cost matrix, $\mathbf{C}$, is then constructed between all feasible pairs based on a weighted sum of the relative pose change and semantic consistency:
\begin{equation}
\mathbf{C}(i,j) = \lambda_1 \norm{\mathbf{p}_{ij}}_2 +\lambda_2 |\phi_{ij}| + \lambda_3 (1-\mathbb{I}[c_i = c_j])
\end{equation}

Object-observation pairs with a geometric dissimilarity, $\epsilon_{ij}$, greater than the similarity threshold, $\theta_{\textrm{sim}}$, not within the association distance threshold, $\theta_{\textrm{dist}}$, or with a cost, $\mathbf{C}(i,j)$, above a cut-off threshold, $\theta_{\textrm{cutoff}}$, are given a cost of infinity. The resulting cost matrix is run through the greedy Hungarian algorithm to find an optimal association. Unassociated observations are added to the mapped object library with an initial state probability distribution (with geometric change expectation, $\mathbb{E}[l]=0$, and stationarity expectation, $\mathbb{E}[v]=v_{\textrm{class}}$), and are marked as \emph{new} objects. For observations that have been assigned to an existing object in the library, the new observation is stored and the object is marked as \emph{update-pending}. Objects with at least $\theta_{\textrm{vis}}$ percent of their points within the current camera frustum, but with no associated observations, are marked as \emph{unobserved}. Note that, since the Hungarian algorithm has a polynomial complexity of $O(n^3)$, we only perform data association within the distance threshold, $\theta_{\textrm{dist}}$, to reduce the computational cost under large scenes. For objects displaced by a large distance, we believe it is not necessary to find the actual correspondence to update the map.

\subsection{Probabilistic Stationarity and Change Update} \label{prob_stat_update}
As \emph{update-pending} and \emph{unobserved} objects are identified, their state distributions are updated via Bayesian inference, as discussed in Section \ref{prob_update}. For an \emph{update-pending} object, $\mathbf{O}_i$, we then estimate the magnitude of geometric change, $\Delta_i\in\mathbb{R}$, as discussed in Section \ref{diff_Scene_change}, to construct the likelihood distribution for the Bayesian update. If $\Delta_i$ is within half of the measurement standard deviation, $\sigma$, which we deem a successful geometric verification, the observation is integrated into the object's TSDF, $\mathbf{M}_i$. Else, the observation is discarded, as it is no longer consistent with the object model from previous observations. For an \emph{unobserved} object, a large pseudo-change is used to penalize its stationarity. Similar to the voxel-level confidence clamping trick used in \cite{hornung13auro}, we do not perform the object-level state update if the new observation brings an object's stationarity score above an upper bound, $v_\textrm{max}$, to ensure responsiveness to changes.

\subsection{Object and Map Update}
\label{obj_map_update}
Once all \emph{update-pending} and \emph{unobserved} objects are updated, the stationarity score expectation, $\mathbb{E}[v_i]$, of each object is checked against a heuristic-based stationarity threshold, $\theta_{\textrm{stat}}$. If the expected score falls below the threshold, all voxels in map that are associated with the object's TSDF, $\mathbf{M}_i$, are reinitialized and the object is erased from the object library, $\mathcal{O}$. All \emph{new} objects are integrated into the library and the map.





\section{Methodology} \label{methodology}

In this section, we present the details of our contributions, including how we estimate changes between object-observation pairs, justifications behind our probabilistic modelling, and the derivation of our Bayesian update rule.

\subsection{Object-Level Geometric Change Estimation} \label{diff_Scene_change}
We first describe our approach to estimate the object-level geometric change using TSDFs. The method is inspired by~\cite{Sucar2020NodeSLAMNO} and~\cite{wang2021dspslam}, where the authors propose to improve 3D reconstruction quality by minimizing the cumulative back-projection depth error between RGB-D images and the TSDF model. Here, we ray-trace from the current camera pose to obtain a scalar-valued, zero-mean error measure between two TSDFs. For each object-observation pair, $\{\mathbf{O}_i,\mathbf{Y}_{t,j}\}$, we first transform the point cloud of the object, $\mathbf{O}_i$, namely $\mathbf{P}_{i}$, and the point cloud of the observation, $\mathbf{Y}_{t,j}$, namely $\hat{\mathbf{P}}_{t,j}$, to the camera optical frame, $\mathcal{C}$. Two TSDFs are constructed around the two point clouds by ray-tracing through the camera optical center, $\mathbf{o}$. We define the change of $\mathbf{O}_i$ with respect to $\mathbf{Y}_{t,j}$ over voxels, $\mathbf{\Omega}$, as:

\begin{equation}
    \Delta_{t,ij} = \textrm{sign}(\mathbf{p}^{\mathcal{C}}_z)   \frac{\lambda_{\textrm{diff}}}{|\mathbf{\Omega}|}\sum_{\mathbf{v} \in \mathbf{\Omega}} | \text{tsdf}(\hat{\mathbf{P}}_{t,j}, \mathbf{v}) - \text{tsdf}(\mathbf{P}_{i}, \mathbf{v})|
\end{equation}
where $\lambda_{\textrm{diff}}$ is a scaling factor and $\mathbf{\Omega}$ is the intersection of the non-zero voxel sets of the two TSDF grids:
$$
\mathbf{\Omega} = \{\mathbf{u} \mid \text{tsdf}(\hat{\mathbf{P}}_{t,j}, \mathbf{u}) > 0 \land \text{tsdf}(\mathbf{P}_{i}, \mathbf{u}) > 0\}.
$$

Here, $\mathbf{p}^{\mathcal{C}}_z$ is the $z$ component of the 3D translation found by the ICP between $\mathbf{P}_{i}$ and $\hat{\mathbf{P}}_{t,j}$ in the camera optical frame, $\mathcal{C}$, where the $z$-axis points forward. Thus, $\mathbf{p}^{\mathcal{C}}_{z}$ is positive if $\hat{\mathbf{P}}_{t,j}$ is in front of $\mathbf{P}_{i}$, and negative otherwise, from the camera's point of view. Therefore, $\Delta_{t,ij}$ takes a positive or negative value depending on the actual geometric change. If the object stays at the same pose, $\Delta_{t,ij}$ is approximately 0. For $unobserved$ objects, we define a large pseudo-change of $\Delta_{t,ij} = \Delta_{\textrm{max}}$ to force the removal of $unobserved$ objects. 

In comparison to other metrics discussed in Section \ref{lit_review}, our approach directly estimates how much the object has changed in the metric space. It utilizes full geometric information of the objects and also takes camera view angle, potential occlusions, and effects of partial observations into account. 

\subsection{Object-Level State and Process Model} \label{obj level state process model}
To support object-level reasoning of the environment dynamics, we introduce a novel probabilistic object state representation. For each object, $\mathbf{O}_i \in \mathcal{O}$, and its previously associated observations, $\{\mathbf{Y}_{t,j}\}_{t=1 \dots T}$, we can extract a sequence of high-level measurement features, $\{\mathbf{z}_{t,j}\}_{t=1 \dots T}$ (see Section~\ref{sec:meas_model}). We would like to jointly estimate the object's geometric change, $l_i \in \mathbb{R}$, and stationarity score, $v_i \in [0,1]$. Dropping the object index, $i$, and observation index, $j$, for clarity, this object-level joint state probability distribution is given by:

\begin{equation}
\label{eq:state_model}
    p\left(l, v \mid \mathbf{z}_{1} \dots \mathbf{z}_{T}\right)
\end{equation}

This joint distribution can be updated iteratively as new measurements arrive, by following a Bayesian update rule:

\begin{equation}
\label{eq:bayesian_update}
\overbrace{p\left(l, v \mid \mathbf{z}_{1} \dots \mathbf{z}_{T}\right)}^{\text{Posterior}}  \propto \overbrace{p\left(\mathbf{z}_T \mid l, v, \mathbf{z}_{1} \dots \mathbf{z}_{T-1}\right)}^{\text{Measurement Likelihood}} \overbrace{p\left(l, v \mid \mathbf{z}_1 \dots \mathbf{z}_{T-1}\right)}^{\text{Prior}}
\end{equation}

Assuming that geometric change, $l$, and stationarity score, $v$, are conditionally independent, we adopt a probability model, first proposed in \cite{vogiatzis2011video} and later verified in \cite{Dong2018PSDFFP,Forster2014SVOFS,Pizzoli2014REMODEPM,Jun} for volumetric fusion and sparse visual SLAM, to parametrize the state distribution in Equation \ref{eq:state_model} as the product of a Gaussian distribution for $l$, and a Beta distribution for $v$:

\begin{equation}
\label{eq:state_param}
\begin{aligned}
    p\left(l, v \mid \mathbf{z}_{1} \dots \mathbf{z}_{T}\right) &\defeq q\left(l, v \mid \mu_T, \sigma_T, \beta_T, \alpha_T\right)\\
        &\defeq \mathcal{N}(l \mid \mu_T, \sigma_T^2)\textrm{Beta}(v \mid \alpha_T, \beta_T)
\end{aligned}
\end{equation}

In all previous works, the Gaussian-Beta parametrization has been used in pixel-level depth estimation and inlier/outlier identification. The same parametrization also fits nicely into our object-level change detection framework, as $l$ is expected to center around an underlying change measure (0 for stationary objects), and $v$ estimates the likelihood of the object being stationary. In Equation \ref{eq:state_param}, $\mu_T$ and $\sigma^2_T$ represent the mean and variance of $l$ respectively, and $\alpha_T$ and $\beta_T$ are the number of observed inlier and outlier measurements with respect to the model (see Section~\ref{sec:meas_model}). Finally, object pruning decisions can be made based on the expectation of the stationarity, $\mathbb{E}[v]$. 
\subsection{Measurement Likelihood Model}
\label{sec:meas_model}
In our setup, each measurement feature, $\mathbf{z}_t$, contains the geometric change measure, $\Delta_{t}$, as calculated in Section~\ref{diff_Scene_change}, and the stationarity class, $s_{t}$, as determined in Section~\ref{3d_inst_Det}: 
\begin{equation}
\mathbf{z}_{t} = \{\Delta_{t}, s_{t} \} \quad \forall t=1 \dots T.
\end{equation}
Applying Bayes rule to the \emph{Measurement Likelihood} term in Equation \ref{eq:bayesian_update}, the measurement likelihood becomes:
\begin{equation}
\label{eq:full_meas}
\begin{array}{l}
    p\left(\mathbf{z}_T \mid l, v, \mathbf{z}_{1} \dots \mathbf{z}_{T-1}\right) \\
    \quad = \hspace{1mm} p\left(\Delta_T,s_T | l, v, \Delta_{1},s_{1} \dots \Delta_{T-1},s_{T-1}\right) \\
    \quad \propto \hspace{1mm} p\left(\Delta_{T} \mid l, v, \Delta_{1}, s_{1} \dots \Delta_{T-1},s_{T-1},s_{T}\right) \\
     \qquad \times p\left(s_{T} \mid l, v, \Delta_1,s_1 \dots \Delta_{T-1},s_{T-1}\right)
\end{array}
\end{equation}

Further, we assume that the geometric terms, $\Delta_t$ and $l$, and the visual-semantic term, $s_t$, are independent, and the measurements, $\mathbf{z}_t$, are conditionally independent given current $l$ and $v$ estimates. Equation \ref{eq:full_meas} can then be simplified as the product between a geometric consistency likelihood, $p\left(\Delta_{T} \mid l, v\right)$, and a stationarity likelihood, $p\left(s_T \mid v\right)$:
\begin{equation}
\label{eq:simp_meas}
\begin{array}{l}
\hspace{5mm} p\left(\Delta_{T},s_{T} \mid l, v, \Delta_{1}, s_{1} \dots \Delta_{T-1}, s_{T-1}\right) \\
\qquad \propto p\left(\Delta_{T} \mid l, v\right) p\left(s_T \mid v\right) 
\end{array}
\end{equation}
We categorize the measurements in one of two ways: as an inlier measurement where the object did not move and the measured geometric change is normally distributed around the current estimate $l$ (initially 0), or as an outlier measurement where the object is moved and the change is uniformly distributed in an interval, $[-\Delta_{\textrm{max}}, \Delta_{\textrm{max}}]$. Inspired by~\cite{vogiatzis2011video}, where the authors use a Gaussian-Uniform mixture to model uncertainties in pixel-level depth measurements, we also model the object-level geometric consistency likelihood as a Gaussian-Uniform mixture weighted by the stationarity score, $v$:
\begin{equation}
\label{eq:depth_meas}
\begin{aligned}
&\hspace{2mm} p\left(\Delta_{T} \mid l, v\right) \defeq v\mathcal{N}(\Delta_T \mid l, \tau^2) \\
& \hspace{2.5cm} + (1-v)\mathcal{U}(\Delta_T \mid -\Delta_{\textrm{max}}, \Delta_{\textrm{max}})
\end{aligned}
\end{equation}
where the measurement variance, $\tau^2$, can be determined experimentally using both static and changed reference objects.

On the other hand, the stationarity class, $s_T=s(c_T)$, can be intuitively considered as a sample from a Bernoulli process controlled by the object's stationarity score, $v$:
%
%
$$
s_T \sim \textrm{Bernoulli}(v)
$$
Note that the Beta stationarity prior of Equation \ref{eq:bayesian_update} can be considered as a conjugate prior to the Bernoulli stationarity likelihood here. As a result, both the geometric consistency measurement, $\Delta_T$, and the stationarity measurement, $s_T$, contribute to the update of the stationarity score, $v$. To balance the relative importance between $\Delta_T$ and $s_T$, we introduce an adaptive factor, $k$, depending on $s_T$ and $\Delta_T$:
%
\begin{equation}
\label{eq:label_meas_final}
\begin{aligned}
&\hspace{5mm} p\left(s_{T} \mid v\right) \defeq  \textrm{Bernoulli}(s_T \mid v)^k
\end{aligned}
\end{equation}

The factor, $k$, acts as a weight in the Beta stationarity update rule for the posterior, as will be shown in Section \ref{prob_update}. It aids in adjusting the model behaviour. For example, the model should adapt quickly when a large geometric change is measured for dynamic objects, such as robots, while being more conservative to static objects, such as shelves. 

\subsection{Probabilistic Update Rule} \label{prob_update}
Combining the measurement models, Equation \ref{eq:depth_meas} and Equation \ref{eq:label_meas_final}, and the parametrized state model, Equation \ref{eq:bayesian_update}, we can derive a Bayesian update rule. If the prior (i.e. the model after processing measurements $\{z_{t}\}_{t=1 \dots T-1}$) is parametrized by $(\mu,\sigma,\alpha,\beta)$, the true posterior would have the form:
\begin{equation}
\label{eq:true_posterior}
\begin{array}{l}
    p(l,v \mid \Delta_T, s_{T}, \mu,\sigma,\alpha,\beta) =\\
    \quad \eta p(\Delta_T \mid l,v)p(s_{T} \mid v)q(l,v \mid \mu,\sigma,\alpha,\beta)
\end{array}
\end{equation}
for some normalization factor, $\eta$. The true posterior can be rearranged to take the form:
\begin{equation}
\label{eq:rearranged_posterior}
\begin{array}{l}
p(l,v \mid \Delta_T, s_{T}, \mu,\sigma,\alpha,\beta) =\\
    \quad C_1 \mathcal{N}(l \mid m, \gamma^2)\textrm{Beta}(v \mid \alpha+ks_T+1,\beta+k(1-s_T)) \\
    \quad + C_2 \mathcal{N}(l \mid \mu, \sigma^2)\textrm{Beta}(v \mid \alpha+ks_T,\beta+k(1-s_T)+1)
\end{array}
\end{equation}
\noindent which is a weighted mixture of two Gaussian-Beta distributions. The weights, $C_1$ and $C_2$, are the probability of the measurement, $\mathbf{z}_T$, being an inlier and outlier, respectively. The first term models the effect of fusing an inlier measurement, where the geometric change, $l$, is updated with a new mean, $m$, and variance, $\gamma^2$. The stationarity score, $v$, also adapts with a $k$-weighted update rule. The second term models the effect of fusing an outlier measurement, where only the stationarity score, $v$, adapts. 

Note that Equation \ref{eq:rearranged_posterior} cannot be written as a single Gaussian-Beta distribution exactly. As in ~\cite{vogiatzis2011video,Dong2018PSDFFP, Jun}, we find a new parametrization for an approximated Gaussian-Beta posterior $q(l,v \mid \mu',\sigma',\alpha',\beta')$ by matching the first and second moments for $l$ and $v$ to the true posterior in Equation~\ref{eq:rearranged_posterior}. The approximated posterior can be found analytically, enabling efficient online inference. We refer the reader to the Supplementary Material for the full derivation.

\section{Experimental Results}
\subsection{Experimental Setup}
We verify the performance of our framework both qualitatively and quantitatively by comparing the map reconstruction of \method~to several baseline methods, each representative of a category of dynamic environment mapping approaches: 
\begin{itemize}
    \item Kimera \cite{Rosinol20icra-Kimera}: Assumes the world is static and adopts a naive TSDF update rule.
    \item MaskFusion \cite{Rnz2018MaskFusionRR}: Actively tracks and updates all potentially dynamic objects.
    \item Fehr \textit{et al.} \cite{fehr2017tsdf}: Performs voxel-level corrections by comparing new observations against the constructed map. 
    \item Panoptic Multi-TSDFs \cite{panoptictsdf}: Maintains objects as static submaps and performs consistency checks to prune those that have changed.
\end{itemize}

\begin{figure}[t!]
  \centering
  \includegraphics[width=\columnwidth]{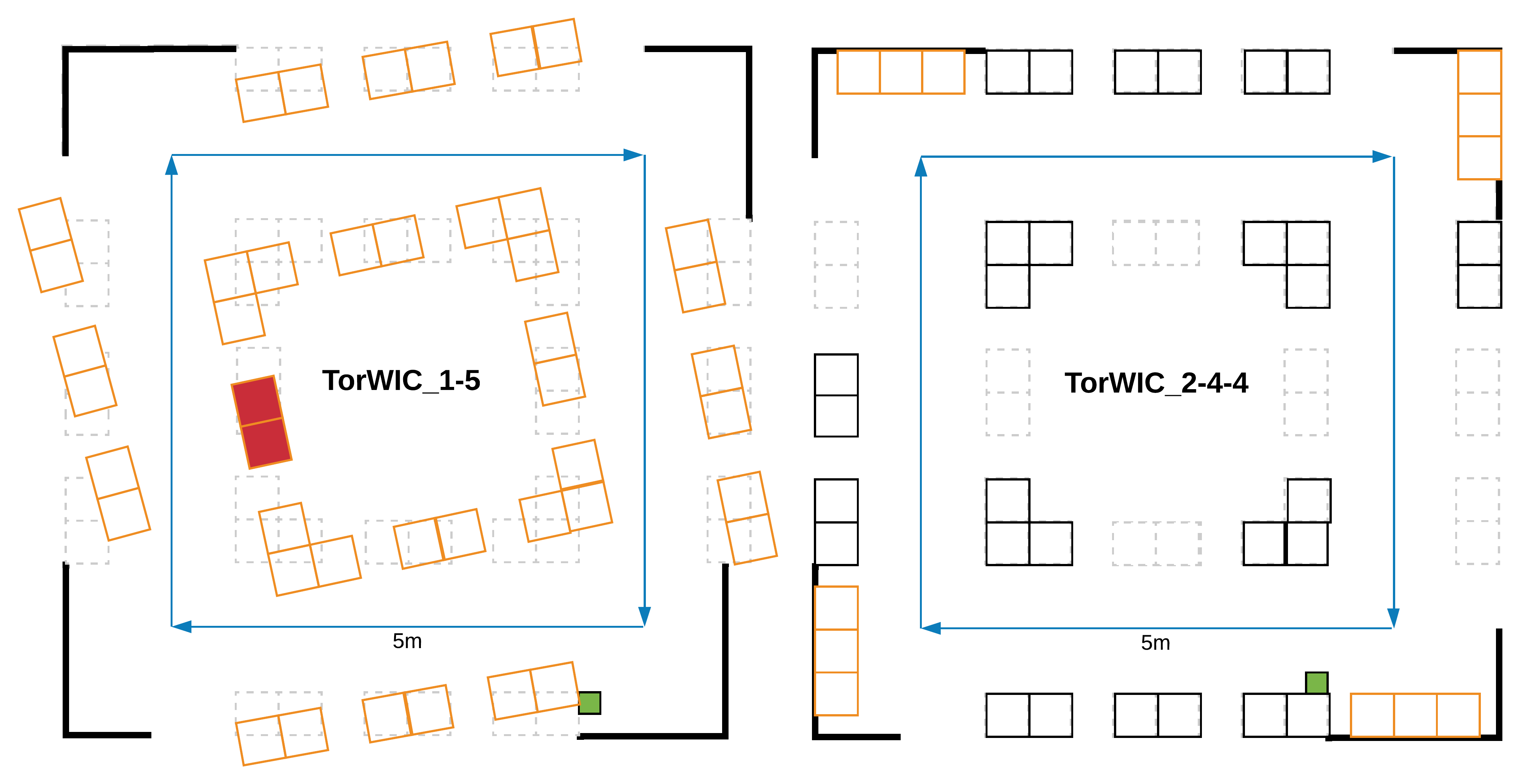}
  \caption{The TorWIC\_1-5 and TorWIC\_2-4-4 ground truth schematics, respectively. Green box: fixed AprilTag for drift reference; black: stationary box or fence; orange: shifted boxes; dotted grey: original position of shifted boxes. The evolution of the red-filled box wall's stationarity score $\mathbb{E}[v]$ on the left is shown in Figure \ref{fig:conf_plot}.}
  \label{fig:clearpath_15}
  \vspace{2mm}
\end{figure}

\begin{figure}[t!]
  \centering
  \includegraphics[width=\columnwidth]{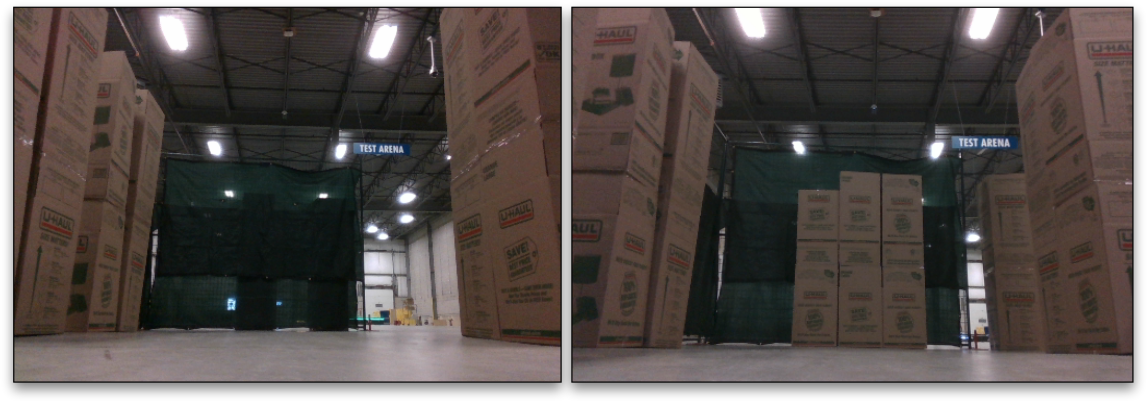}
  \caption{An RGB image comparison of the robot revisiting the same spot (AprilTag) during the TorWIC 2\_4\_4 route. Changes include 3 stacks of boxes added in front of the fence, and an additional box wall to the right of the fence. The ground truth schematic of this changed route can be seen in Figure \ref{fig:clearpath_15}, right.}
  \label{fig:compare_2-4}
\end{figure}

\begin{figure*}[ht!]
  \centering
  \includegraphics[scale=0.65]{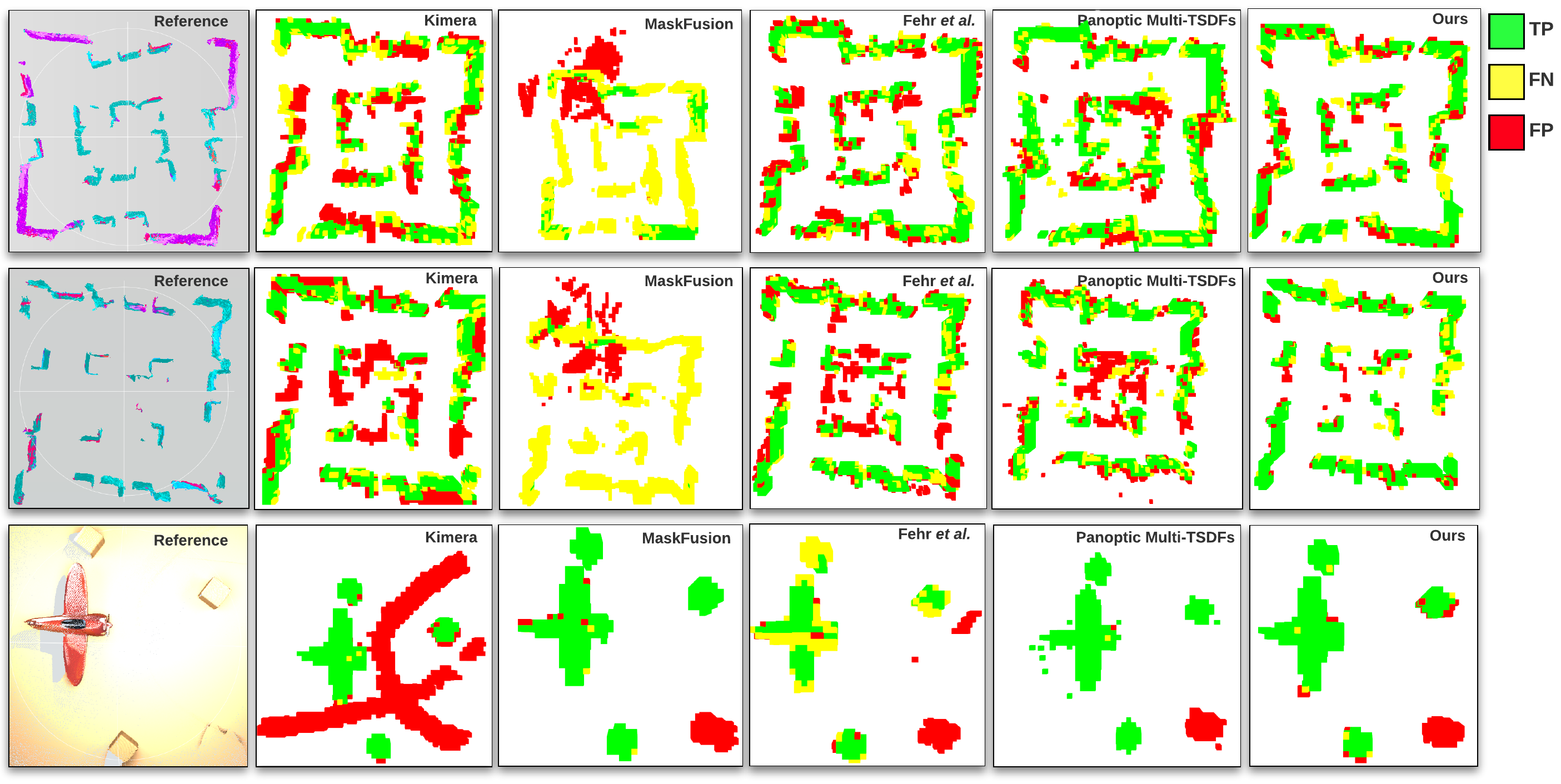}
  \caption{Bird's-eye-view qualitative 3D reconstruction results of the \textbf{top row:} TorWIC\_1-5 route, \textbf{middle row:}  TorWIC\_2-4-4 route, and \textbf{bottom row:}  ToyCar scenario. The reconstruction produced by \method~is compared against that of Kimera, MaskFusion, Fehr \textit{et al.}, and Panoptic Multi-TSDFs. The green, yellow, and red sections represent true positives, false negatives, and false positives, respectively. The first image is the reference map of the routes' final scenario. A voxel size of 20 cm for the TorWIC routes and 6 cm for the ToyCar scenario is used to compute the metrics.}
  \label{fig:results_15}
\end{figure*}

To demonstrate the capabilities of \method, we evaluate on two datasets. Due to the lack of real-world semi-static datasets, we create one for the purpose of this problem (Section~\ref{dataset}). Furthermore, we use the ToyCar dataset from~\cite{cofusion} to show that our framework not only handles semi-static changes, but is also robust to dynamics. As the work of Fehr \textit{et al.} is not open-source, it was implemented on top of Voxblox~\cite{voxblox} based on~\cite{fehr2017tsdf}. We fine-tune the parameters of the baselines on our dataset whenever possible for the best reconstruction. We implement \method~on top of Kimera~\cite{Rosinol20icra-Kimera}.

\subsection{The TorWIC Dataset} \label{dataset}

We release TorWIC, a real-world warehouse dataset collected on a OTTO 100  Autonomous Mobile Robot equipped with a RealSense D435i RGB-D camera, a 2D Hokuyo UAM501 LiDAR, a wheel encoder, and an inertial measurement unit (IMU). The robot setup can be seen in Figure 3, the sensor specifications are listed in Table 3, and sample sensor data can be seen in Figure 4, of the Supplementary Material. 

We collected 18 real-world trajectories with 4 types of changes, each with a unique warehouse-like scenario, containing the following objects: 
\begin{itemize}
    \item 2m-tall walls made out of cardboard boxes
    \item 3m-tall tarp-covered fences at the end of each hallway
\end{itemize}
\urldef{\trajsdoc}\url{https://github.com/Viky397/TorWICDataset/blob/main/TorWIC_Dataset.pdf}
The robot makes multiple passes of each of the 18 scenarios to accumulate sufficient sensor coverage. Reference poses are obtained by running a proprietary LiDAR-based SLAM algorithm for each trajectory. The scenarios differ by the removal, addition, shift, or rotation of boxes or fences. Since each trajectory starts and ends in the same spot, identifiable by the AprilTag, they can be stitched in different ways using the provided script to create longer routes that present changed object locations over time. Figure~\ref{fig:clearpath_15} shows the scenario changes for two sample routes, TorWIC\_1-5 (trajectory 1\_1 and 1\_5 stitched) and TorWIC\_2-4-4 (trajectory 2\_2, 2\_4, and 4\_1 stitched), and Figure \ref{fig:compare_2-4} shows two RGB images captured by the robot along the TorWIC\_2-4-4 route. A high level breakdown of the trajectories can be found in Table~4 of the Supplementary Material. A document with detailed descriptions of each scenario is provided\footnote{\trajsdoc}.

\subsection{Real-World Experiment in a Semi-Static Environment}

We first perform qualitative and quantitative evaluations against the four baseline methods on two routes from the TorWIC dataset, as shown in Figure \ref{fig:clearpath_15}. Since a ground truth map is not available, and each method reconstructs the scene differently, we create a set of reference maps by processing the final scenario through each method. For a qualitative evaluation, we visually inspect the reconstruction against the corresponding reference map. For a quantitative evaluation, we overlay and voxelize both the reference and reconstructed maps and count overlapping and inconsistent voxels to compute the precision, recall, and false positive rate (FPR) for each method at the voxel level. We list the parameters used in \method ~in the Supplementary Material.

The top two rows of Figure \ref{fig:results_15} contain the bird's-eye-views of the reference and reconstructions of the two TorWIC routes. Quantitative evaluation results are listed in Table \ref{tab:clrpath_15}. When exporting the final map, we downsample and threshold voxel weights to prune the low confidence regions such that all methods produce a similar recall. Here, precision showcases the framework's coverage of true objects, and FPR showcases how well objects are updated after a scene change. Therefore, the precision and FPR should be compared for performance evaluation. It can be seen that \method~produces the most visually accurate map, with minimal ghost artifacts or double walling. It also produces the highest precision and lowest FPR. \\ 
\indent When compared to the baselines, Kimera does not handle shifted objects, leaving many artifacts such as double walls in the reconstruction. MaskFusion relies on both geometric and semantic consistency to track objects. However, when faced with partial observations of visually similar boxes, it fails to track the box wall due to ambiguous associations and inaccurate object motion estimates, resulting in an incorrectly constructed map with a low precision and recall, despite being given the reference robot poses. The method of Fehr \textit{et al.} is able to partially correct the map for regions inside the camera's field-of-view (FOV), but is still left with artifacts and ghosts outside the camera's FOV, due to the absence of an object-level understanding of the environment. Panoptic Multi-TSDFs is also able to partially correct the map. In contrast to the method of Fehr \textit{et al.}, it maintains object-level information, thus is able to perform consistent map updates for partially visible objects. However, this method lacks an explicit object association module and detects changes by counting inconsistent voxels over overlapping submaps. As a consequence, it behaves similarly to the naive TSDF update when an object has moved too far from its initial position and no association is established. 

\begin{table}[t!]
\tiny
\centering
\caption{Quantitative mapping results on the TorWIC dataset.}
\resizebox{\columnwidth}{!}{
\begin{tabular}{c|ccc}
\Xhline{2\arrayrulewidth}  \Xhline{2\arrayrulewidth}
\textbf{TorWIC\_1-5} & \textbf{Precision $\uparrow$} & \textbf{Recall (TPR) $\uparrow$} & \textbf{FPR $\downarrow$} \\ \hline
Kimera         & 60.6   &  74.2  & 7.7   \\
\rowcolor{LightCyan}
MaskFusion    &  54.5  &  24.4 & 3.5 \\
Fehr \textit{et al.}     & 76.3 & 76.6   & 3.8    \\
\rowcolor{LightCyan}
Panoptic Multi-TSDFs &  77.4  &  \textbf{78.9}  &  4.2  \\
\hline 
\textbf{\method~(ours)}    &  \textbf{80.2}  & 78.7  & \textbf{3.0}  \\ 
\textit{\method~Improvement}      & 2.8 & 	-0.2 & 	0.5 \\\hline
\Xhline{2\arrayrulewidth} \Xhline{2\arrayrulewidth}
\textbf{TorWIC\_2-4-4} & \textbf{Precision $\uparrow$} & \textbf{Recall (TPR) $\uparrow$} & \textbf{FPR $\downarrow$} \\ \hline
Kimera         & 55.8    &   78.3  & 8.5    \\
\rowcolor{LightCyan}
MaskFusion       &  23.8  & 5.5   & 3.0 \\
Fehr \textit{et al.}     &  67.4  &  80.7  &   5.3  \\
\rowcolor{LightCyan}
Panoptic Multi-TSDFs  &  62.7  &  77.1  &   5.3  \\
\hline
\textbf{\method ~(ours)}    &  \textbf{87.0}  &  \textbf{81.2}  &  \textbf{1.6} \\ 
\textit{\method~Improvement}  & 19.6 & 	0.5 & 	1.4 \\\hline
\end{tabular}}
\label{tab:clrpath_15}
\end{table}

\begin{figure}[t!]
  \centering
  \includegraphics[width=0.95\columnwidth]{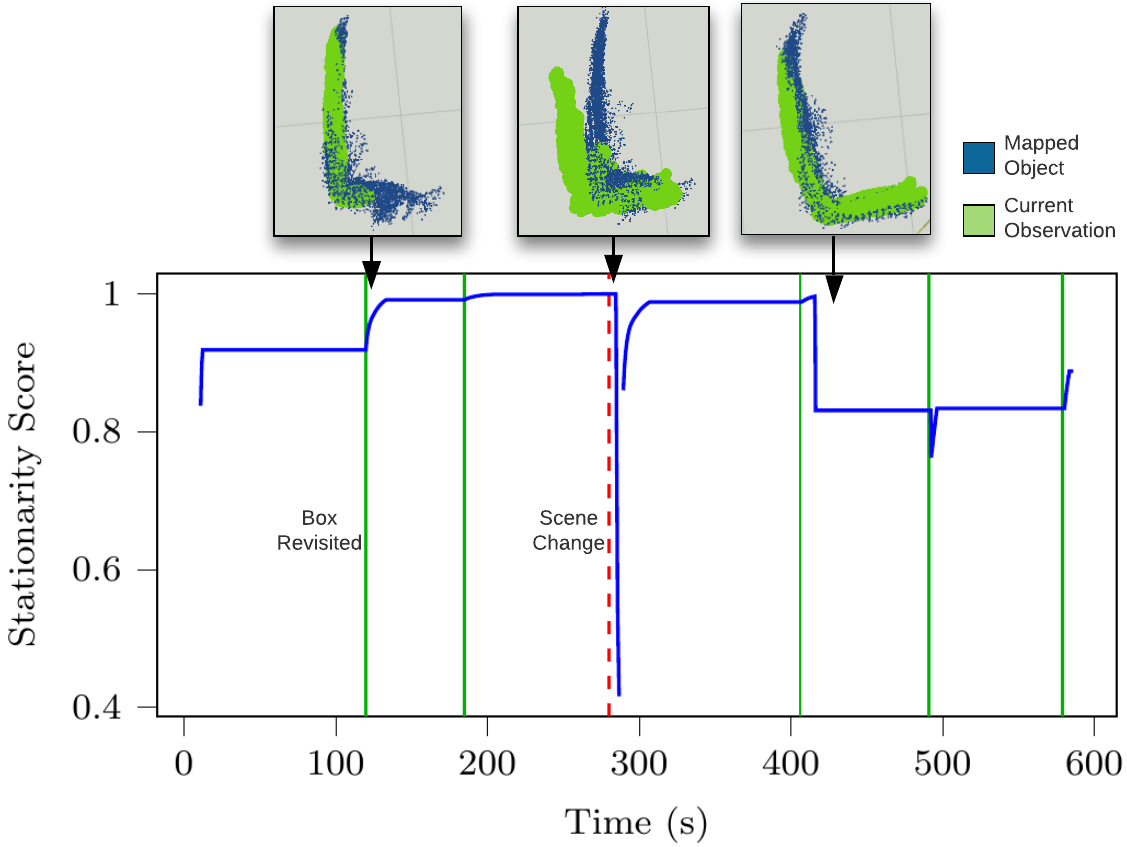} 
  \caption{Evolution of the expected stationarity, $\mathbb{E}[v]$, of the red-filled boxes in Figure \ref{fig:clearpath_15}, as the robot makes multiple loops. The stationarity score adapts each time the box is revisited along the robot's trajectory, indicated by the green lines. The expectation increases or decreases depending on how well the new observations match the mapped object model, with the mapped object and current observation visible at three separate times. The object is discarded and reconstructed when the scene changes, indicated by the red dotted line.}
  \label{fig:conf_plot} 
\end{figure}

\begin{figure}[b!]
  \centering
  \includegraphics[width=\columnwidth]{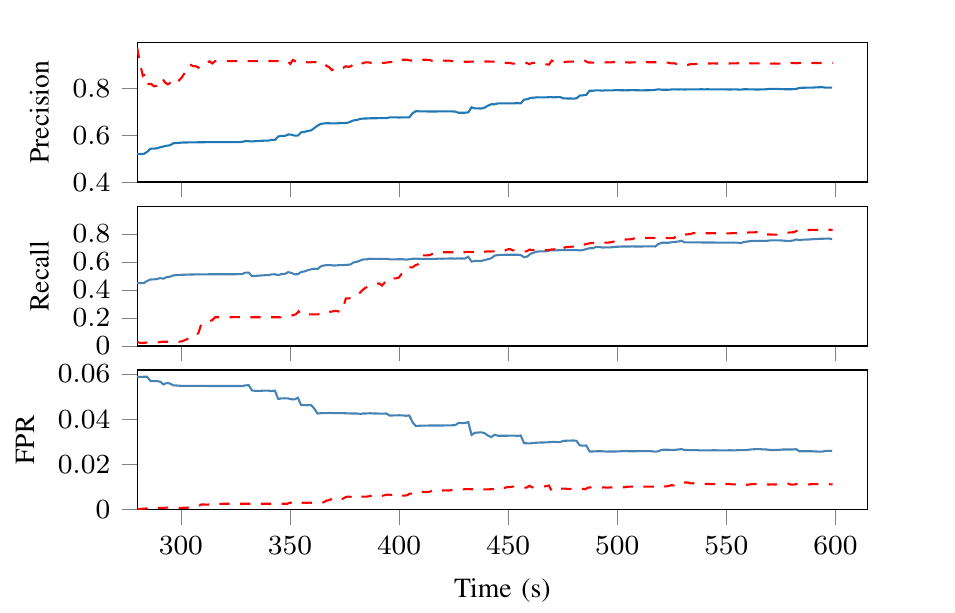}
  \caption{Evolution of the precision, recall, and false positive rate (FPR) of \method~after the scene has changed, on the TorWIC\_1-5 trajectory, starting from a map of the previous scenario (solid blue line), and from scratch (dashed red line).}
  \label{fig:fpr_plot}
\end{figure}

As discussed in Section \ref{prob_stat_update}, \method~propagates a belief distribution for the stationarity score, $v$, for each object in the scene as the robot accumulates more measurements. Figure~\ref{fig:conf_plot} shows the evolution of the stationarity score's expectation for a sample box wall. As the robot revisits the box wall, its estimate of the object being stationary adapts. As the scene changes, its stationarity drops below the threshold, $\theta_{\textrm{stat}}$, and is erased and recreated at its new location. The drop in stationarity seen around 400 seconds is due to inaccuracies in poses obtained during dataset collection. However, the measurement is still considered an inlier by the measurement model and thus integrated into the object's reconstruction, increasing the stationarity score after more observations are made. 

Finally, we study \method's~effectiveness on map maintenance after a scene change. Figure~\ref{fig:fpr_plot} plots the precision, recall, and FPR of the reconstructed map over time. We perform two runs with \method~after the scene change: one continued run starting with a prior map constructed before the scene change, and another reference run without any prior maps. The reference run reconstructs the post-change scene from scratch and should produce the highest quality map achievable for all shifted objects. As seen in Figure~\ref{fig:fpr_plot}, \method~is able to steadily increase both precision and recall, and reduce FPR, as the robot moves through the environment. The precision starts off low for the continued run as the percentage of change in the environment is large. Therefore, the box walls reconstructed in the prior map no longer reflect the new scenario. However, a large portion of the voxels in the prior map, such as those belonging to the unchanged fences, are still valid, leading to a high recall at the start and throughout the run. The final map has a comparable recall to that of the reference, with a slightly lower precision and slightly higher FPR due to regions not seen during the second trajectory. This evaluation shows how our framework effectively corrects inconsistencies in the map when sufficient observations have been made.

\subsection{Robustness Against Dynamics}

We will now evaluate the methods' robustness to dynamic objects, specifically in comparison to the works of Fehr \textit{et al.}, and Panoptic Multi-TSDFs, which also focus on mapping in semi-static scenes. For this experiment, we use the simulated ToyCar dataset from \cite{cofusion}, as they provide a ground truth camera trajectory, depth, and instance-level segmentation images. Similar to the real-world experiment, we construct a set of reference background maps using each method, by masking out the two dynamic toy cars. We tune parameters for each method to produce the best visually reconstructed map, while maintaining a similar recall for a fair comparison. 

The third row of Figure \ref{fig:results_15} contains the bird's-eye-view of both reference and reconstructed scenes of the ToyCar dataset, with the respective quantitative analyses listed in Table \ref{tab:tc}. Kimera does not handle dynamics in the environment, leaving a trail behind the two toy cars as they drive. In contrast to the real-world experiment, MaskFusion is able to produce a highly accurate scene reconstruction as it is now provided with instance-level segmentation, thus enabling accurate motion estimation of the toy cars. The work of Fehr \textit{et al.} is able to produce a clean reconstruction as well, though it does not operate at an object-level. Therefore, voxels that belong to complex surfaces are pruned aggressively, resulting in poor object completeness and a low precision and recall. Moreover, artifacts are left behind the toy cars when they are blocked from the camera. Panoptic Multi-TSDFs achieves the most accurate final reconstruction. \method~produces a clean final reconstruction comparable to that of MaskFusion and Panoptic Multi-TSDFs. Note that the toy car is masked as false positive at its final position (bottom right corner) in all reconstructions, as it is not present in the reference background map.
\begin{table}[t!]
\tiny
\centering
\caption{Quantitative mapping results on the ToyCar dataset.}
\resizebox{\columnwidth}{!}{
\begin{tabular}{c|ccc}
\Xhline{2\arrayrulewidth}  \Xhline{2\arrayrulewidth}
\textbf{ToyCar} & \textbf{Precision $\uparrow$} & \textbf{Recall (TPR) $\uparrow$} & \textbf{FPR $\downarrow$} \\ \hline
Kimera         & 24.9   &   97.4  & 13.4   \\
\rowcolor{LightCyan}
MaskFusion       &   80.2  & 97.8   & 1.8 \\
Fehr \textit{et al.}     &  62.0  & 51.3  &   2.0 \\
\rowcolor{LightCyan}
Panoptic Multi-TSDFs  &  \textbf{82.0}   & \textbf{99.2}  &  \textbf{1.2} \\
\hline
\textbf{\method ~(ours)}    &  79.4  &  96.9  &  1.9 \\ 
\textit{POCD Improvement}      &  -2.6 & -2.3 & 0.7 \\ \hline
\end{tabular}}
\label{tab:tc}
\vspace{-5.0mm}
\end{table}

\begin{figure}[t!]
  \centering
  \includegraphics[width=0.85\columnwidth]{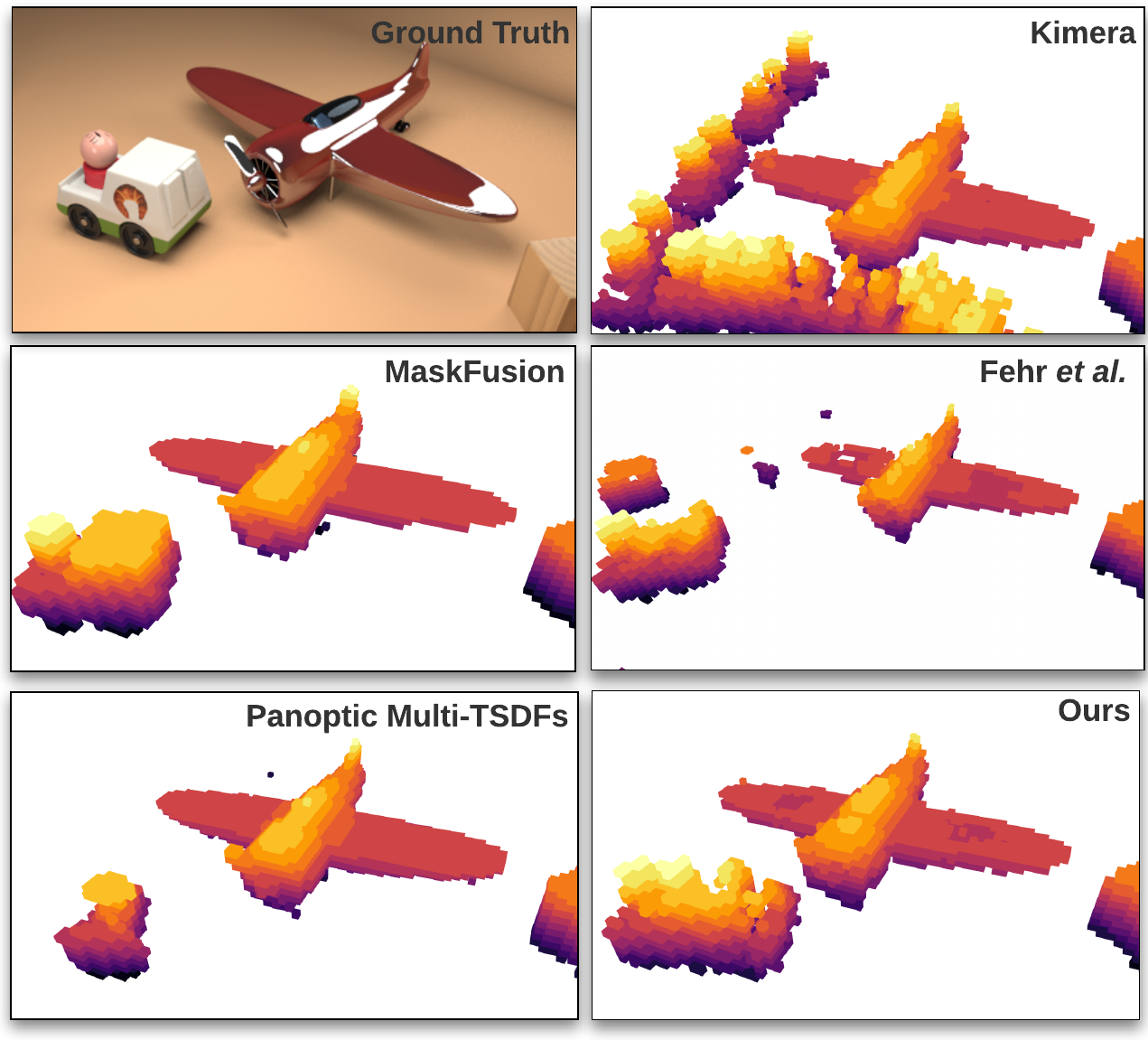}
  \caption{Qualitative comparison of the toy car reconstruction during its motion. Note that Kimera does not handle any dynamics whereas MaskFusion is designed for dynamic scenes.}
  \label{fig:4cars}
\end{figure}

While the methods that attempt to handle changes (MaskFusion, Fehr \textit{et al.}, Panoptic Multi-TSDFs, \method) achieve comparable reconstructions of the final scene, their reconstruction quality for the dynamic toy cars varies based on the assumptions they adopt. In Figure \ref{fig:4cars}, we compare the reconstruction quality of one of the toy cars during the run. Kimera assumes a static scene, thus failing to handle the toy car. MaskFusion is specifically designed to handle dynamic objects, producing the most visually accurate model and a smooth trajectory. The other three methods target semi-static environments, and do not explicitly handle dynamic objects: all these methods erase and reconstruct the toy car at its new position after sufficient observations are gathered. The work of Fehr \textit{et al.} is able to correctly construct and update this toy car as it is never occluded. Though, as map updates are performed at a voxel-level, this approach performs ineffectively under occlusions, as can be seen by the resulting trail left behind the other toy car shown in the middle-right row. \\
\indent Panoptic Multi-TSDFs assumes all objects are static and does not perform explicit object-level association. Hence, when the toy car moves out of its original bounding region, a correct association is not established and the object-level map update degrades to the traditional voxel-level TSDF update behaviour. As seen in the bottom-left row of Figure \ref{fig:4cars}, the resulting reconstruction is eroded when the toy car leaves its submap. Alternatively, \method~features an object-level association module allowing it to achieve correct associations even in the presence of dynamics. In contrast, \method~can robustly update the toy car at an object-level, even when it has moved a large distance. Note that as we focus on semi-static changes in this work, \method~is not able to perform incremental object pose tracking for dynamic objects, as updates to object poses only occur when sufficient confidence is achieved that the object has moved. We plan to explore explicit handling of dynamics as an extension to this work.

\subsection{Runtime Analysis}
We evaluate the performance of our framework on a Linux system with an AMD Ryzen R9-5900X CPU with 24 threads at 3.7GHz. Our runtime analysis does not account for segmentation, which is performed offline, as it is not the focus of this work. The average runtime performance is measured on the TorWIC\_1-5 route over all frames. With an input resolution of 640$\times$360 and a voxel size of 5 cm, the average per-frame computation time for point clustering and object-observation association, object state update and model integration, and map maintenance is 118.30 ms, 49.80 ms and 50.72 ms, respectively. Together, we achieve an average FPS of 4.57, compared to 22.50 FPS for the original Kimera Semantics codebase, which our system builds on, and 5-6 FPS for Panoptic Multi-TSDFs \cite{panoptictsdf}. The most expensive operation is clustering and association, varying our frame rate from 1.25 FPS when 43 objects are mapped, to 34.24 FPS when only two objects are mapped. Some box walls are mapped as multiple objects due to a limited sensor FOV. \method~is amenable to online operation as map maintenance is not required at every frame. Note that our codebase is not optimized for computation time, and we expect to achieve significantly higher performance by utilizing parallel processing and GPUs.

\section{Conclusion}\label{sec:conclusion}
In this paper, we present \method, a novel online, object-aware map maintenance framework that simultaneously tracks and reconstructs incremental changes in the environment by leveraging both geometric and semantic information. \method~performs Bayesian updates to propagate a joint state distribution for each tracked object in the scene. We experimentally verify the robustness of \method~against several state-of-the-art baselines on two datasets, including TorWIC, a real-world warehouse dataset that we release with this work. The results suggest that object-level knowledge of the environment, explicit object tracking, consistent object-level change estimation, and different levels of dynamics-handling are all crucial for reliable long-term map maintenance.



\bibliographystyle{unsrtnat}
\bibliography{references}

\end{document}